\documentclass[10pt, conference, compsocconf]{IEEEtran}

\usepackage{graphicx}
\usepackage{lipsum}
\usepackage{float}
\usepackage{amsmath}
\usepackage{subcaption}
\usepackage{caption}
\usepackage{color}
\usepackage{hyperref}

\begin{document}
	%
	\title{\vspace{-5ex}A Reflectance Based Method \\ For Shadow Detection and Removal \vspace{-4ex}}

	
	\author{
		\IEEEauthorblockN{Sri Kalyan Yarlagadda, Fengqing Zhu}
		\IEEEauthorblockA{School of Electrical and Computer Engineering, Purdue University, West Lafayette, Indiana, USA.\vspace{-4ex}}
		
	}
	
	\maketitle
		
	\begin{abstract}
		Shadows are common aspect of images and when left undetected can hinder scene understanding and visual processing. We propose a simple yet effective approach based on reflectance to detect shadows from single image. An image is first segmented and based on the reflectance, illumination and texture characteristics, segments pairs are identified as shadow and non-shadow pairs. The proposed method is tested on two publicly available and widely used datasets. Our method achieves higher accuracy in detecting shadows compared to previous reported methods despite requiring fewer parameters. We also show results of shadow-free images by relighting the pixels in the detected shadow regions. 
		
	\end{abstract}
	
	\begin{IEEEkeywords}
		shadow detection, reflectance classifier, shadow removal, image enhancement.
	\end{IEEEkeywords}

	%

	\section{Introduction}
Shadows are ubiquitous. They are formed when light is partially or fully occluded by objects. Shadows provide information about lighting direction \cite{Efros2009}, scene geometry and scene understanding \cite{Wehrwein2015} in images and are crucial for tracking objects \cite{Gao2016} in videos. They also form an integral part of aerial images \cite{Shedlovska2016}. However, shadows can also complicate tasks such as object detection, feature extraction and scene parsing \cite{Zhu2010}. 
	
	There have been many methods proposed to detect shadows from images and videos \cite{Gao2016,Lalonde2010,Zhu2010,Khan2015,Guo2013,Chung2009,Qi2012}. In this paper we focus on detecting shadows from color images.  With the recent boom in data driven approaches, machine learning based methods have been applied to detect shadows \cite{Zhu2010,Khan2015,Guo2013}. In \cite{Zhu2010} Conditional Random Fields consisting of 2490 parameters are used to detect shadows in gray scale images using features such as intensity, skewness, texture, gradient similarity etc. In \cite{Khan2015} Convolutional Neural Networks consisting of 1000's of parameters are used to detect shadows. In \cite{Lalonde2010} intensity information around edges is used to detect shadow boundaries. In \cite{Guo2013} image is first segmented and various classifiers are used to detect regions similar in color and texture by comparing different segments with each other. 
	
	\begin{figure}[h]
		\centering
		\includegraphics[scale = .15]{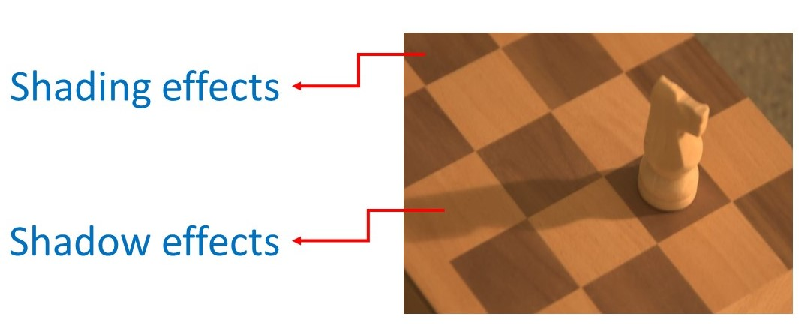}
		\caption{Note that both the surfaces are dark but one is due to shadow and another is due to shading. Such examples complicate shadow detection but can be solved using neighborhood information}
		\label{fig:comp}
	\end{figure}
	
	In this paper we propose a non-training based shadow detection method which requires fewer parameters, yet achieves high accuracy compared to previous methods \cite{Guo2013,Zhu2010,Khan2015}. We differ from \cite{Guo2013} in the features and classifiers used for comparing regions and also in the approach of using these comparisons to obtain the shadow mask. Every surface is characterized by two features: its reflectance and its texture. When a shadow is cast on a surface its illuminance reduces, but its reflectance remains the same. Due to the reduction in the illuminance, there will also be some loss in texture information. By examining a surface, it is difficult to tell whether it is dark due to the effects of shadow or shading. An example of this is given in Figure \ref{fig:comp}. By comparing surfaces with each other we can detect shadows with greater confidence. Hence, by pairing different regions of an image based on their reflectance, texture and illumination characteristics we can detect shadows efficiently. 
	
	\section{Shadow Detection}
	\label{sec:method}
\begin{figure}[h]
		\centering
		\begin{subfigure}{.2\textwidth}
			\centering
			\includegraphics[width=1\linewidth]{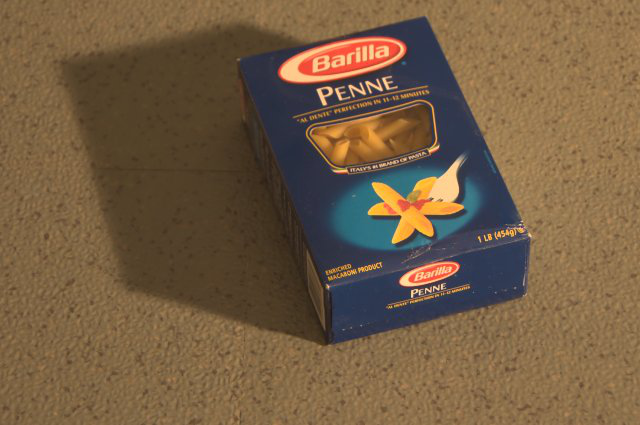}
			\caption{Original Image}
			\label{fig:seg-sub1}
		\end{subfigure}
		\begin{subfigure}{.2\textwidth}
			\centering
			\includegraphics[width=1\linewidth]{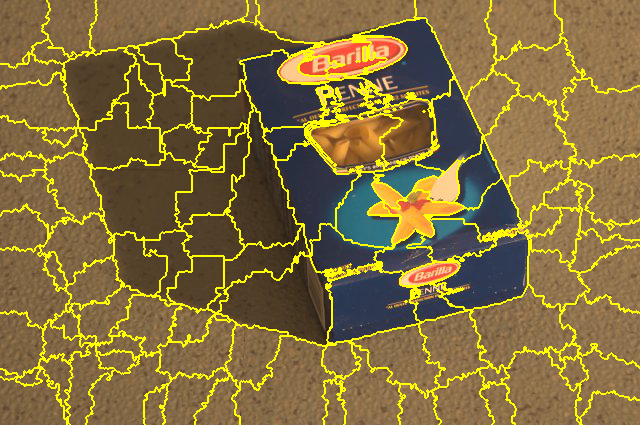}
			\caption{Segmented Image}
			\label{fig:seg-sub2}
		\end{subfigure}
		\caption{An example of a test image segmented using Quickshift with a kernel size of 9. The segmentation correctly separates the boundaries between the shadow and non-shadow regions.}
		\label{fig:seg}
	\end{figure}

	Our goal is to group different regions of an image based on their reflectance, texture and illumination characteristics. 
	To group pixels with similar properties into different regions, we first segment an image using the Quickshift method \cite{vedaldi08quick} with a Gaussian kernel size of 9. Our assumption is that a single segment should contain pixels with similar reflectance and illumination. An example of segmentation result is shown in Figure \ref{fig:seg}. In the subsections below we explain how we design the reflectance, texture and illumination classifiers to label each segment as shadow or non-shadow.
	
	\subsection{Reflectance classifier}    
\begin{figure}[t]
		\centering
		\begin{subfigure}{.2\textwidth}
			\centering
			\includegraphics[width=1\linewidth]{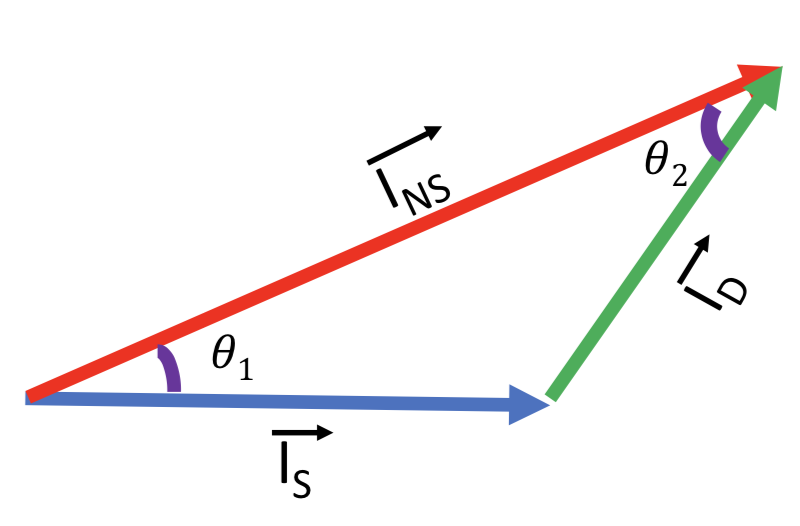}
			\caption{}
			\label{fig:angle-sub1}
		\end{subfigure}
		\begin{subfigure}{.2\textwidth}
			\centering
			\includegraphics[width=1\linewidth]{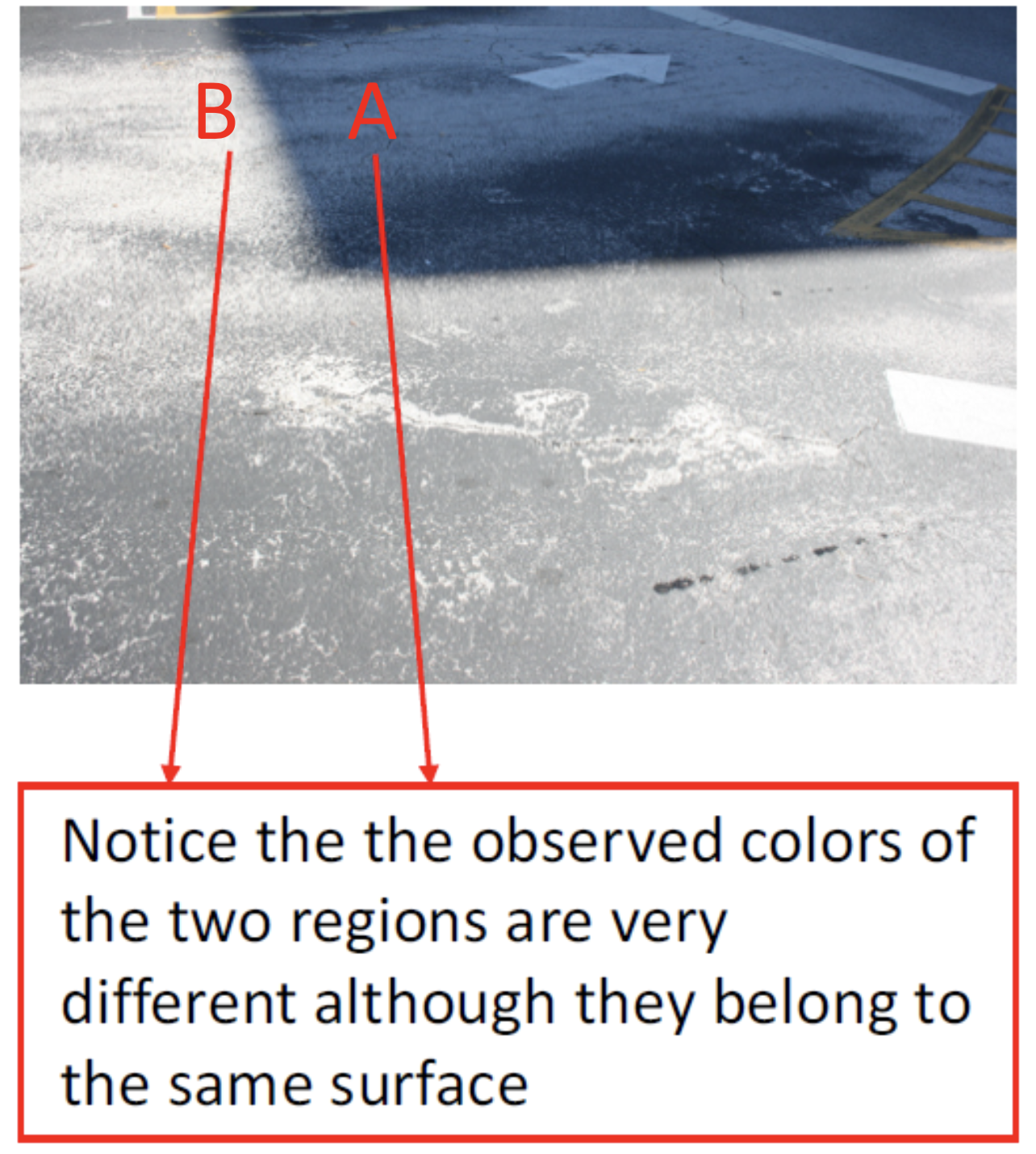}
			\caption{}
			\label{fig:angle-sub2}
		\end{subfigure}
		\caption{An illustration of color vectors is given in \ref{fig:angle-sub1}. In \ref{fig:angle-sub2} visually region A (the shadow region) on the road appears blue, not gray. This is due to the large chromaticity difference in direct light $L_d$ and the reflected light $L_e$ leading to a large $\theta_1$. However, the angle between $I_D$ and $I_{NS}$ ($\theta_2$) will be small so that we can classify region A and region B as shadow non-shadow pairs.}
		\label{fig:angle}
	\end{figure}

	Consider the illumination model used in \cite{Guo2013},
	
	\begin{equation} 
	I_i = (t_icos(\theta)L_d + L_e )R_i 
	\end{equation} 
	
where $I_i$ is the vector representing the $ i^{th} $ pixel in RGB space, $L_d$ and $L_e$ are vectors representing the direct light and reflected light from the environment, respectively. $\theta$ is the angle between direct light and surface normal and $R_i$ is the reflectance vector. The value of $t_i$ indicates whether the pixel belongs to shadow or non-shadow segment. When $t_i = 0$ the pixel is in the shadow segment and vice versa. Two segments belonging to the same surface but under different illumination can be modeled as $t_i = 0 $ for all the pixels in the shadow segment and  $t_i = 1$ for all the pixels in the non-shadow segment. Assuming that direct light and environment light are constant in magnitude and direction over the two segments, we can see that the reflectance property of the surface remains constant in both cases. Taking the respective median of all pixels in RBG color space in each of the segments, we have the following, $$I_{NS} - \mbox{Median color of non-shadow segment}$$ $$I_{S} - \mbox{Median color of shadow segment}$$ $$I_{D} = I_{NS} - I_{S} = (cos(\theta)L_d)R^{median}$$  In the case when $L_{e}$ is similar to $L_d$ in terms of chromaticity, the angle between the vectors $I_{D}$ and $I_{NS}$ should be zero. However, in practice $L_e$ differs from $L_d$, hence these two vectors will have a small angle provided they are of the same material and a large angle if they are of different material with different reflectance properties. By thresholding the angle between the color vectors $I_{D}$ and $I_{NS}$, we can decide whether two segments with different illumination conditions belong to the same material. We call this the ``angle criterion." Notice that we don't look at the angle between $I_{NS}$ and $I_{S}$ because in the case where $L_e$ is significantly different from $L_d$ in terms of chromaticity, the angle between these two vectors will be very large even if they represent the same surface. We set the angle threshold to be $10^{\circ}$. An example of this case is illustrated in Figure \ref{fig:angle}.

	\subsection{Luminance Classifier}
	Shadows are formed when direct light is partially or fully occluded and hence have lower illumination. The decrease in illumination depends on the relative intensities of $L_d$ and $L_e$. A large decrease in illumination intensity darkens the shadow. To build an effective luminance classifier, we need to be able to detect the decrease in illumination and be able to attribute that decrease to obstruction of light and not due to some noise. In order to model this, we look at the luminance values of all pixels in the LAB color space. We compute the median luminance of all segments in the LAB space and compute the histogram of the median luminance values. The peaks of the histogram give us an estimate of the number of different illumination regions in the image.
	
\begin{figure}[t]
		\centering
		\begin{subfigure}{.2\textwidth}
			\centering
			\includegraphics[width=1\linewidth]{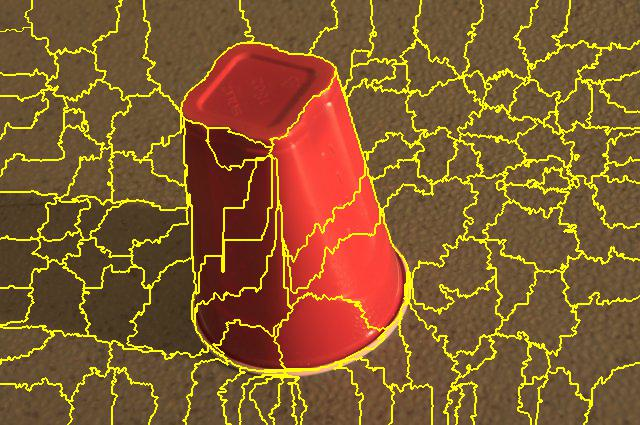}
			\caption{Segmentation}
			\label{fig:luminance-sub1}
		\end{subfigure}
		\begin{subfigure}{.2\textwidth}
			\centering
			\includegraphics[width=1\linewidth]{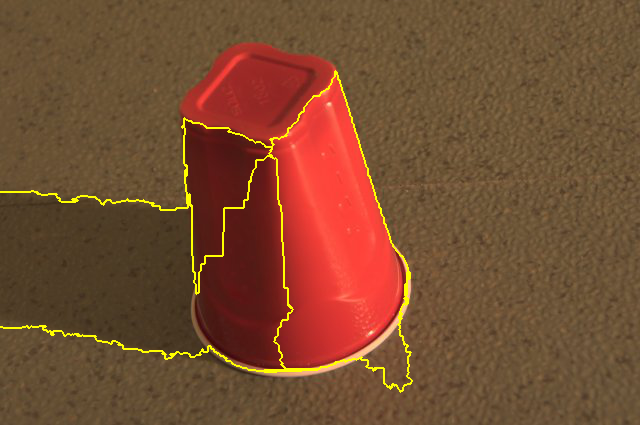}
			\caption{Luminance clustering}
			\label{fig:luminance-sub2}
		\end{subfigure}
		\caption{The segmented image in \ref{fig:luminance-sub1} is grouped using the luminance classifier and the result is shown in \ref{fig:luminance-sub2}.}
		\label{fig:luminance}
	\end{figure}

	We then split the image into regions by grouping segments based on their proximity to the peaks. Segments within the same region are not compared because they have similar illumination intensity while segments from different groups are allowed for comparison to detect shadows. This step is useful because it adaptively groups segments into regions with similar illumination. An example of grouping segments into regions based on their luminance is shown in Figure \ref{fig:luminance}. In addition to the grouping criteria, for two segments to be shadow non-shadow pairs, the ratio of their median luminance $T$ in LAB space has to be above the threshold of 1.2 in order to avoid comparing segments with similar illumination. $T$ can be anywhere between 1 and $\infty$ and the closer it is to 1 the closer the illumination intensities of the two segments are. Shadow non-shadow pairs will have a high values of $T$ compared to segments with similar illumination intensities. 
	
	\subsection{Texture classifier}
	
	Since shadow and the corresponding non-shadow segments are of the same material their texture characteristics will be similar. However, due to the reduction in illumination intensity of shadow segments, some texture information is lost. To capture this phenomenon, we look for texture similarity between the segments under comparison provided that their $T$ is not very high, because if its high a lot of texture information would have been lost. We compute the Earth Mover Distance  between the histograms of the texton maps \cite{Malik} of both segments and threshold it to find whether the two segments have similar texture. However, if $T$ is greater than 2.4 we do not compare them for texture similarity as a lot of texture information is lost in the shadow segment due to the decrease in illumination. 
		
	\subsection{Implementation}
\begin{figure}[h]
		\centering
		\begin{subfigure}{.2\textwidth}
			\centering
			\includegraphics[width=1\linewidth]{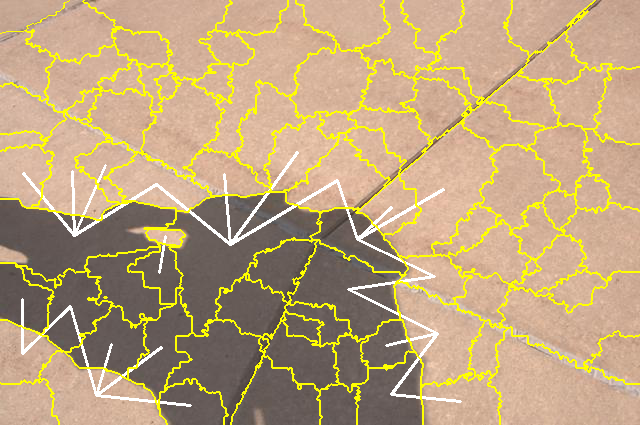}
			\caption{First Iteration}
			\label{fig:neighbor-sub1}
		\end{subfigure}
		\begin{subfigure}{.2\textwidth}
			\centering
			\includegraphics[width=1\linewidth]{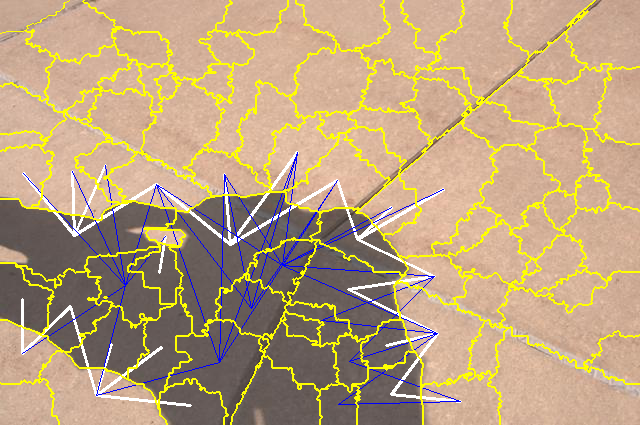}
			\caption{Second Iteration}
			\label{fig:neighbor-sub2}
		\end{subfigure}
		\caption{The connections obtained by first iteration are marked with the white lines (\ref{fig:neighbor-sub1}) and the connections obtained by second iteration are marked by the blue lines (\ref{fig:neighbor-sub2}).}
		\label{fig:neighbor}
	\end{figure}

	In this subsection, we describe how we use the above three classifiers to detect shadow non-shadow segment pairs. Each segment is compared to its neighboring segments using the reflectance, texture and luminance classifiers discussed above. If all the classifiers label the pair as a shadow non-shadow pair, we store that connection. We use these connections to connect more segments. For every shadow non-shadow pair, we take all the non-classified neighbors of the shadow segment and compare them to non-shadow segment using the above classifiers. We repeat this process 3 times. The reason is that some shadow segments may have neighbors which are also shadow segments themselves. Such segments will not be detected in the first iteration. In order to connect them to the already labeled shadow segments, we repeat the process by using the information obtained from the initial connections. The process is illustrated in Figure \ref{fig:neighbor}.
	
	\subsection{Refinement}
	\label{subsec:refinement}
	The above implementation detects shadow non-shadow pairs with similar reflectance, texture and different luminance but does not put any constraint on how bright the shadow segment should be. Without such constraint, two very bright segments can be misclassified as shadow non-shadow pairs. In order to avoid this, we limit the shadow region to have a gray scale value lower than the Otsu threshold of the image. We segment the image again with a Gaussian kernel of size 3 (smaller than the Guassian kernel used in the initial segmentation) and look for segments which contain shadow pixels using the initial shadow mask. A finer segmentation mask leads to better modeling of the shadow non-shadow boundaries. Given a segment contains shadow pixels, if more than $70\%$ of pixels in that segment have a gray scale value less than the Otsu threshold, we label the entire segment as a shadow segment and if not we label the entire segment as non-shadow. 
	
	\section{Experimental Results}
	The proposed method is evaluated on two publicly available datasets, the UIUC dataset \cite{Guo2013} and the UCF dataset \cite{Zhu2010}. 
	
	\subsection{UIUC Dataset}
	The UIUC dataset consists of 108 images with shadows, out of which 32 images have been used for training and 76 for testing by \cite{Guo2013}. We have evaluated our method on the 76 test images. In addition to computing the per class accuracy, we also show the Balanced Error Rate (BER) for our method which is computed as the following, 
	\begin{equation}
		\text{BER} = 1 - \frac{1}{2}(\frac{TP}{TP+FN} + \frac{TN}{TN+FP})
		\end{equation}
		where FP is False Positives, FN is False Negatives, TP is True Positives and TN is True Negatives. The lower the BER the better the method. BER is used because there are fewer shadow pixels than non-shadow pixels in the images. The results of our methods and others are shown in table \ref{tab:uiuc}. We have achieved the highest accuracy detecting shadows and a very close BER compared to \cite{Khan2015} which has the smallest BER of all three methods.
	
	\begin{center}
		\captionof{table}{Results Our Proposed Method Compared to Other Methods On UIUC Dataset} 		 
		\begin{tabular}{|p{2cm} | c | c | c|}
			\hline
			Methods &	Shadows & Non-Shadows & BER\\ 
			\hline
			Unary + Pairwise(\cite{Guo2013}) & .716 & .952 & .166 \\
			\hline
			ConvNet(\cite{Khan2015}) & .847 & .955  & \textbf{.099} \\ 
			\hline
			Our method  & \textbf{.906} & .855 &  .119 \\
			\hline
		\end{tabular}
		\label{tab:uiuc}
	\end{center}
	
	\subsection{UCF Dataset}
	The UCF dataset is also widely used for testing shadow detection methods. It consists of 355 images which are more diverse and complex than the UIUC dataset. In \cite{Zhu2010} 120 images were used for testing. We have tested our method on 236 images. Out of the 236 images, for 162 of them we have followed the proposed method, but for 74 images from OIRDS \cite{oircds} dataset we have chosen a threshold of .35 instead of using the Otsu threshold for limiting the gray scale of the shadow pixels. This is because OIRDS dataset contains aerial images with very dark shadow regions. The results are reported in Table \ref{tab:ucf-our} and comparisons to other methods are shown in Table \ref{tab:ucf-compare}. In comparison to other methods, our method achieved the highest accuracy in detecting shadows and also has the best BER.
	
	\begin{center}
		\captionof{table}{Detection Confusion Matrices of Our Proposed Method On UCF Dataset}
		\begin{tabular}{|p{3cm}|c|c|}
			\hline
			\textbf{74} images from \textbf{OIRDS} dataset & Shadow & Non Shadow \\
			\hline
			Shadow & $.899$ & $.101$ \\
			\hline
			Non - Shadow & $.116$ & $.884$ \\
			\hline
			\hline
			\textbf{162} images from UCF dataset & Shadow & Non Shadow \\
			\hline
			Shadow & $.922$ & $.078$ \\
			\hline
			Non - Shadow & $.191$ & $.809$ \\
			\hline
		\end{tabular}
		\label{tab:ucf-our}
	\end{center}
	
	\begin{center}
		\captionof{table}{Detection Confusion Matrices of Our Proposed Method Compared to Other Methods On UCF Dataset}
		\begin{tabular}{|p{2cm} | c | c | c |}
			\hline
			Methods &	Shadows & Non-Shadows & BER \\ 
			\hline
			BDT-BCRF\cite {Zhu2010} & .639 & .934 & .2135 \\
			\hline
			Unary + Pairwise(\cite{Guo2013}) & .733 & .937 & .165 \\
			\hline
			ConvNet(\cite{Khan2015}) & .780 & .926 & .147 \\ 
			\hline
			Our method & \textbf{.920} & .827 & \textbf{.1265}  \\
			\hline
		\end{tabular}
		\label{tab:ucf-compare}
	\end{center}
	
	\section{Shadow Removal}
	
\begin{figure}[h]
		\centering
		\begin{subfigure}{.2\textwidth}
			\centering
			\includegraphics[width=1\linewidth]{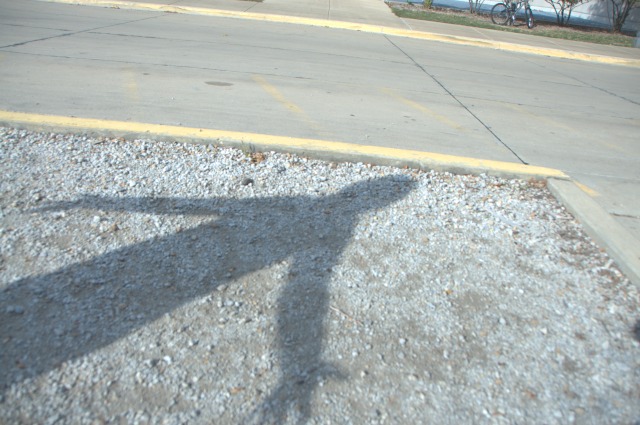}
			\caption{Original Image}
			\label{fig:removal-sub1}
		\end{subfigure}
		\begin{subfigure}{.2\textwidth}
			\centering
			\includegraphics[width=1\linewidth]{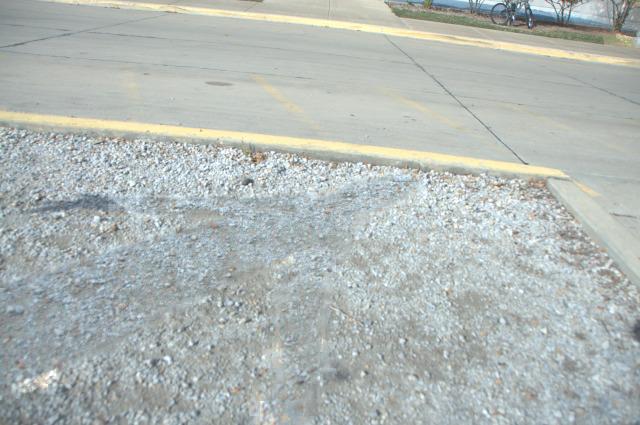}
			\caption{After shadow removal}
			\label{fig:removal-sub2}
		\end{subfigure}
		\begin{subfigure}{.2\textwidth}
			\centering
			\includegraphics[width=1\linewidth]{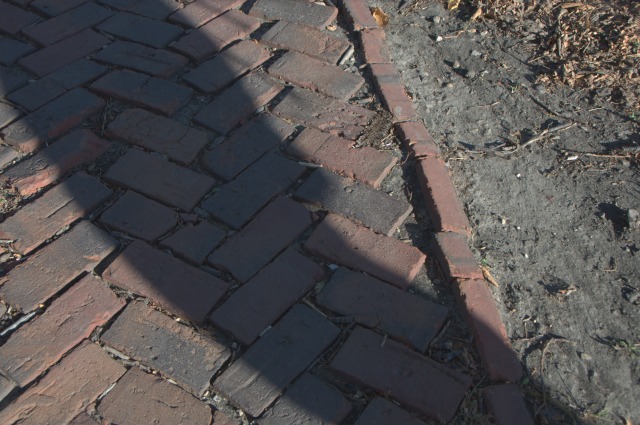}
			\caption{Original Image}
			\label{fig:removal-sub3}
		\end{subfigure}
		\begin{subfigure}{.2\textwidth}
			\centering
			\includegraphics[width=1\linewidth]{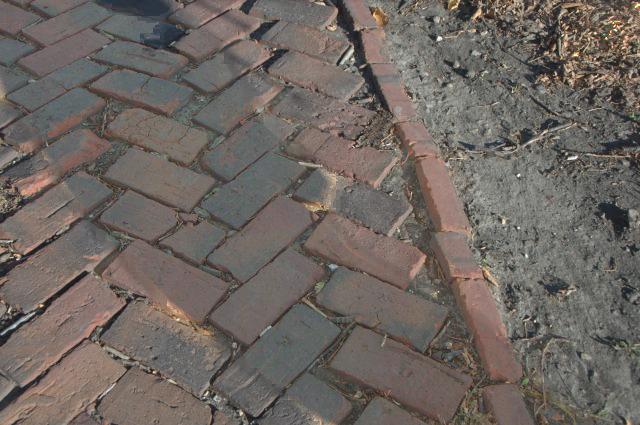}
			\caption{After shadow removal}
			\label{fig:removal-sub4}
		\end{subfigure}
		\caption{Sample shadow removal results.}
		\label{fig:removal}
	\end{figure}

	To remove shadows we follow the same approach as described in \cite{Guo2013}. Some examples of shadow removal are shown in Figure \ref{fig:removal}.
	
	\section{Conclusion}
	We proposed a simple yet effective shadow detection method requiring few parameters. Each image was first segmented and segment pairs were identified as shadow non-shadow pairs based on their reflectance, illumination and texture characteristics. Experimental results showed that our method was effective for detecting shadows but had a lower accuracy in identifying non-shadows. The connections between the detected shadow and non-shadow pairs were used to successfully remove shadows in test images. 
	


	
	
	%

	\bibliographystyle{IEEEtran}
	\bibliography{ms}
	
\end{document}